\documentclass[10pt,twocolumn,letterpaper]{article}

\usepackage{cvpr}
\usepackage{times}
\usepackage{epsfig}
\usepackage{graphicx}
\usepackage{amsmath}
\usepackage{amssymb}
\usepackage{lib}
\usepackage{booktabs}
\usepackage{enumitem}

\usepackage[numbers,sort]{natbib}
\usepackage[font=small,labelfont=bf]{caption}

\usepackage{multirow}
\usepackage{color}
\usepackage[pagebackref=true,breaklinks=true,letterpaper=true,colorlinks,bookmarks=false]{hyperref}

\cvprfinalcopy 

\def\cvprPaperID{3095} 

\ifcvprfinal\fi
\begin{document}

\title{Use the Force, Luke! \\
Learning to Predict Physical Forces by Simulating Effects
}

\author{Kiana Ehsani$^{*2}$, Shubham Tulsiani$^1$, Saurabh Gupta$^3$, Ali Farhadi$^2$, Abhinav Gupta$^{1,4}$\\
$^1$ FAIR, $^2$ University of Washington,
$^3$ UIUC, $^4$ Carnegie Mellon University  \\
\url{https://ehsanik.github.io/forcecvpr2020}
}

\twocolumn[{%
\renewcommand\twocolumn[1][]{#1}%
\vspace{-3em}
\maketitle
\vspace{-2em}
\begin{center}
   \centering \includegraphics[width=\textwidth]{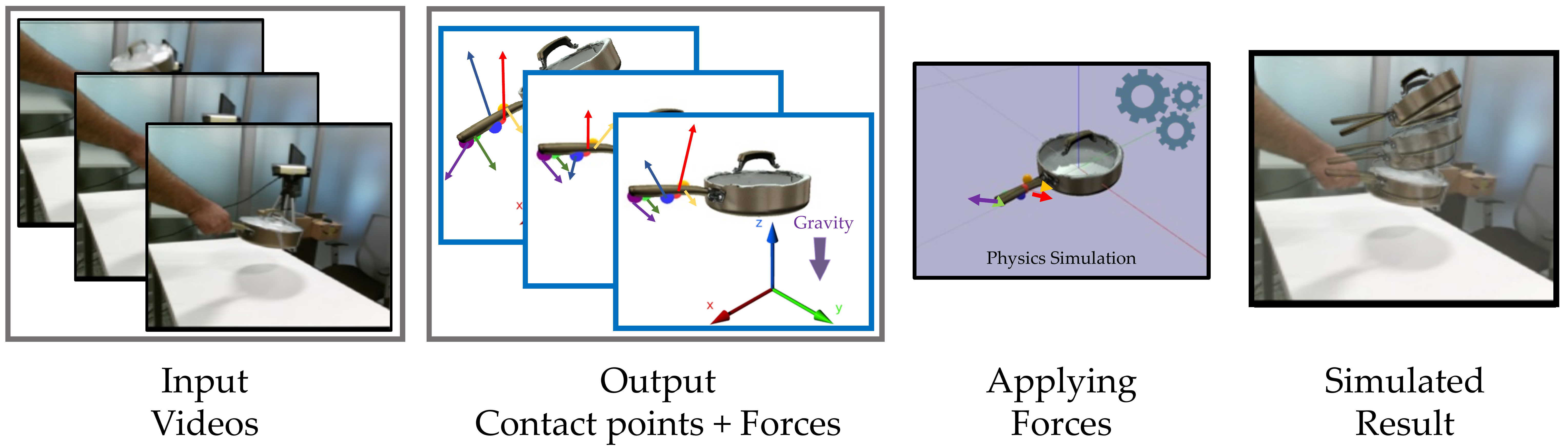} \captionof{figure}{We study the problem of physical understanding of human-object interactions by inferring object motion, points of contacts and forces from RGB videos. We use physics simulation to see how these interactions change the dynamics of the object and try to imitate the motion observed in the video.}
   \label{fig:teaser}
\end{center}%
}]
\thispagestyle{empty}

\let\thefootnote\relax\footnotetext{$^*$ Work done during an internship at FAIR.}

\begin{abstract}
When we humans look at a video of human-object interaction, we can not only infer what is happening but we can even extract actionable information and imitate those interactions. On the other hand, current recognition or geometric approaches lack the physicality of action representation. In this paper, we take a step towards more physical understanding of actions. We address the problem of inferring contact points and the physical forces from videos of humans interacting with objects. One of the main challenges in tackling this problem is obtaining ground-truth labels for forces. We sidestep this problem by instead using a physics simulator for supervision. Specifically, we use a simulator to predict effects, and enforce that estimated forces must lead to same effect as depicted in the video. Our quantitative and qualitative results show that (a) we can predict meaningful forces from videos whose effects lead to accurate imitation of the motions observed, (b) by jointly optimizing for contact point and force prediction, we can improve the performance on both tasks in comparison to independent training, and (c) we can learn a representation from this model that generalizes to novel objects using few shot examples.
\end{abstract}

\section{Introduction}

What does it mean to understand a video of human-object interaction such as the one shown in Figure \ref{fig:teaser}? One popular answer would be to recognize the nouns (objects) and verbs (actions) -- e.g., in this case lifting a pot. But such an understanding is quite limited in nature. For example, simply recognizing `lifting' does not tell one anything about how the pot was grasped, or how high it was lifted. To address these shortcomings, there has been a recent push towards a deeper geometric understanding of videos. From estimating contact points on the object~\cite{brahmbhatt2019contactdb} to estimating human and object poses~\cite{li2019estimating}, these approaches tend to estimate the visible geometric structure. 
While the high-level semantic labeling ('lifting') or the geometric inferences (human and object pose estimation), both provide an answer to \emph{what} happened in the video, it lacks the true physical substance for actionable understanding. For example, just knowing how the pot is moved is not sufficient for the robot to imitate -- it needs to also understand \emph{how} the act was accomplished.

In order to obtain a more actionable understanding, we argue that one must account for the physical nature of the task. The answer then, to the question of how the act was done, is rather straightforward from a physical perspective -- the object was in contact with the human hand on two sides, and a combination of inward and upward forces applied at these contact points allowed it to be lifted up against gravity.  
This understanding of physical \emph{forces} is not only directly useful for an active agent but also completely represents the interaction from the object’s perspective as only external forces cause its motion. 
In this work, we take a step towards developing such an understanding and present a system that can infer the contact points and forces applied to a known object from an interaction video. 

While the goal of being able to infer these forces is desirable, it is unfortunately tedious (if not impossible) to acquire direct supervision for this task. The existing force sensors~\cite{liu2008design} are not precise enough to provide accurate direction or magnitude measurement, and not compact enough to keep the interaction natural. 
So, how do we get the supervision? We note that if we can infer physical forces applied to an object, we can also recover a full geometric understanding by simulating the effect of the forces on that object. We build on this insight and present an approach to learn prediction of physical forces, that instead of directly supervising the predicted forces, enforces that their effects match the observations through the interaction video.

To train our system, we collect a dataset of videos recorded from multiple participants grabbing and moving objects. Then we use Mechanical Turk to annotate the keypoints of the objects and contact points on each frame and use this limited information in camera frame to infer the object's 6DOF pose in world coordinates and the person's contact points on object mesh. We observe that our approach of learning to predict forces via supervising their effects allows us to learn meaningful estimates and that these can explain the observed interaction in terms of reproducing the observed motion. 

Our experiments show that our model learns to infer human contact points on object mesh, and estimate the corresponding forces. We observe that applying these forces on the predicted contact points at each time step in physics simulation we can repeat the behavior depicted in the video. We also show that contact point and force prediction are highly correlated and jointly optimizing improves the performance on both tasks. Finally, we provide interesting evidence that the representation we learn encodes rich geometric and physical understanding that enables us to generalize to interacting with novel objects using only few shot examples.

\section{Related Work}

\vspace{1mm}
\noindent \textbf{Pose estimation.} In order to understand physical motions, a network needs to implicitly reason about the object pose. There is a long line of work in this area, with two different approaches: category-based ~\cite{gupta2015aligning, sahin2018category, song2014sliding} and instance-based~\cite{doumanoglou2016recovering, kehl2016deep, tremblay2018deep, xiang2017posecnn}. 
Our work is more aligned with the latter, and we use the YCB object set~\cite{ycb} which provides richly textured objects. 
While some of our model design decisions is inspired by these works, \eg iterative pose estimation~\cite{li2018deepim}, our final goal, to infer about physics of the observed motions, is different.

\vspace{1mm}
\noindent \textbf{Contact point prediction.} Predicting the contact point and hand pose estimation for object manipulation has been studied in the domain of detecting plausible tool grasping~\cite{akizuki2018tactile}, human action recognition~\cite{garcia2018first}, hand tracking ~\cite{hamer2010object}, and common grasp pattern recognition~\cite{huang2015we}.  Brahmbhatt \etal~\cite{brahmbhatt2019contactdb} collected a dataset of detailed contact maps using a thermal camera, and introduced a model to predict diverse contact patterns from object shape. 
While our model also  reasons about the contact points,it is only one of the components towards better understanding the physical actions. Moreover, we show that we benefit from force prediction to improve the contact point estimations.

\vspace{1mm}
\noindent \textbf{Human Object interaction.} 
Typical approaches for understanding human object interaction use high level semantic labels~\cite{gkioxari2018detecting, gupta2015visual, yatskar2016situation}. Recently there have been some works in understanding the physical aspects of the interaction.
Pham \etal~\cite{pham2017hand} have used data from force and motion sensors to reason about human object interactions. They also use off-the-shelf tracking, pose estimation, and analytical calculations to infer forces~\cite{pham2015towards}. Hwang \etal~\cite{hwang2017inferring} studied the forces applied to deformable objects and the changes it makes on object shape. Li \etal~\cite{li2019estimating} reasoned about human body pose and forces on the joints when the person is interacting with rigid stick-like hand tools. While these methods show encouraging results for this direction, our work concentrates on interaction scenarios with complex object meshes and more diverse contact point patterns.

\vspace{1mm}
\noindent \textbf{Predicting physical motions.}
Recently, learning the physics dynamic has been widely studied by classifying the dynamics of objects in static images~\cite{mottaghi2016newtonian}, applying external forces in synthetic environments~\cite{mottaghi2016happens}, predicting post-bounces of a ball~\cite{purushwalkam2018bounce}, simulating billiard games~\cite{fragkiadakipredictive}, and using generative models to produce plausible human-object interactions~\cite{wang2019learning}.
These works are more broadly related to understanding the physical environment, however, their goal to predict how scenes evolve in the future is different from ours. We try to tackle the problem of physically reasoning about the motions observed in the videos.

\vspace{1mm}
\noindent \textbf{Recovering physical properties.} 
In recent years there have been efforts in building differentiable physics simulation~\cite{de2018end, hu2019chainqueen}. Wu \etal~\cite{wu2015galileo} use physics engines to estimate physical properties of objects from visual inputs. However, in contrast to these approaches aimed at retrieving the properties of the physical world, we assume these are known and examine the problem of interacting with it. Our real-to-sim method is more aligned with the path taken by ~\cite{chebotar2019closing}; that being said, our goal is not bringing the simulation's and real world's distributions closer. We rather focus on replicating the observed trajectory in simulation.
\section{Approach}

Given a video depicting a human interacting with an object, our goal is to infer the physical forces applied over the course of the interaction. This is an extremely challenging task in the most general form. For example, the object geometry may vary wildly and even alter over the interaction (\eg picking a cloth), or forms of contact may be challenging (\eg from elbowing a door to playing a guitar). We therefore restrict the setup to make the task tractable, and assume that the interaction is with a known rigid object (given 3D model), and only involves a single hand (five fingers apply the force). Given such an interaction video, our goal is then to infer the forces applied to it at each time-step along with the corresponding contact points.

Formally, given a sequence of images $\{I_t\}$ depicting an interaction with a known object, and additional annotation for (approximate) initial object pose, we predict the person's contact points in object coordinates frame $C_t=(c^0_t, \dots, c^4_t)$ (each representing one finger), and the forces applied to each contact point $(f^0_t, \dots, f^4_t)$. As alluded to earlier, it is not possible to acquire supervision in the form of the ground-truth forces over a set of training interactions. Our key insight is that we can enable learning despite this, by instead acquiring indirect supervisory signal via enforcing that the simulated effect of predicted forces matches the observed motions across the video. We first describe in \secref{dataset} a dataset we collect to allow learning using this procedure. We then describe in \secref{forcepred} how we can extract supervisory signal from this data, and finally present our overall learning procedure in \secref{learning}.

\subsection{Interaction Dataset}
\seclabel{dataset}
We collect a dataset of object manipulation videos, representing a diverse set of objects, motions, and grasping variations. We leverage the objects from the YCB set~\cite{ycb}, as these have the underlying geometry available, and record a set of videos showing participants manipulating 8 objects. To enable learning using these videos, we collect additional annotations in the form of semantic keypoints and pixel locations of contact points to (indirectly) allow recovering the motion of the objects as well as the 3D contact points for each interaction.

\begin{figure*}[tp]
    \centering
     \includegraphics[width=37pc]{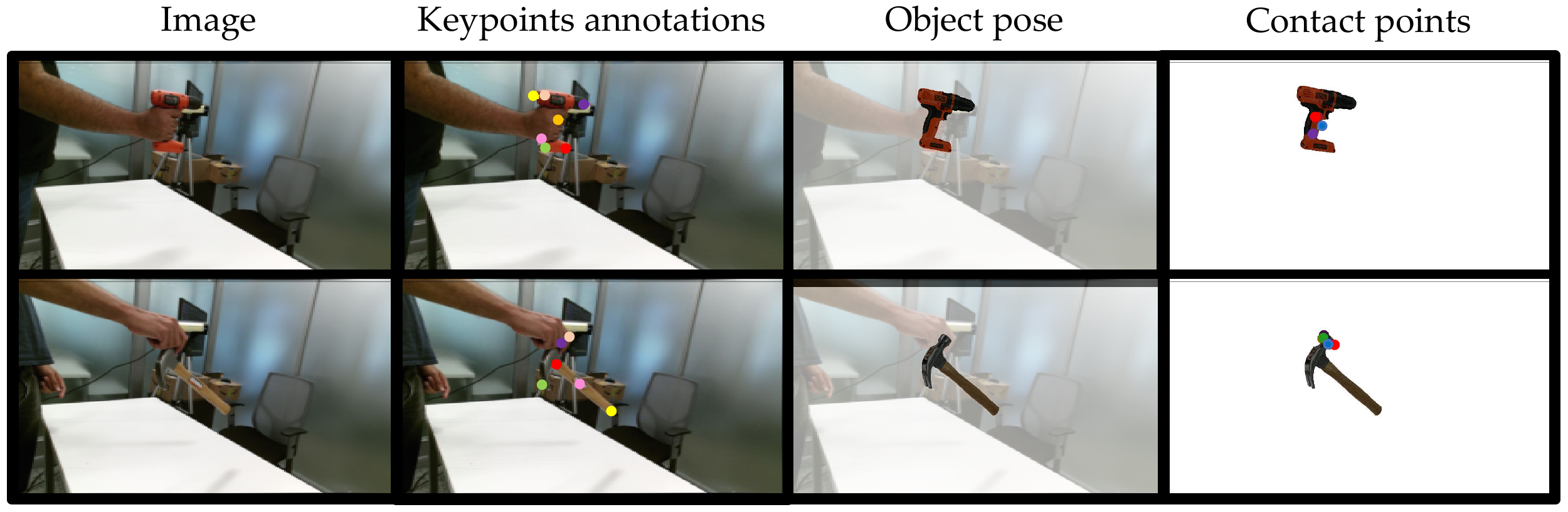}
    \caption{\textbf{Dataset.} Showing two sample frames (drill and hammer) from our dataset. We collect annotation for semantic keypoints of the objects (Column 2) and human contact points. This data helps us to calculate object 6DOF pose (Column 3) and contact points on object mesh (Column 4).}
    \label{fig:dataset}
    \vspace{-1.5em}
\end{figure*}

We describe the data collection procedure in more detail in the supplementary, but in summary, we  obtain: a) annotations for 2D locations for visible keypoints in each frame, b) 6D pose of the object w.r.t. the camera in each frame, though these are noisy due to partial visibility, co-planar keypoints, \etc, and c) 3D contact points on the object mesh over each interaction video. There are 174 distinct interaction videos in our dataset  (111 train, 31 Test, 32 Validation), 13K frames in total. The 8 objects used are: pitcher, bleach bottle, skillet, drill, hammer, toy airplane, tomato soup can and mustard bottle. We show some examples from the dataset in Figure~\ref{fig:dataset}. We will publicly release the dataset and believe that it will also encourage future research on understanding  physical interactions.

\subsection{Supervisory Signal via Physical Simulation}
\seclabel{forcepred}
Given per-timestep predicted forces $f_t$ and corresponding contact points $C_t$, we show that these can get supervisory signal by simulating their effects, and comparing the simulated motion against the observed one. 

\vspace{1mm}
\noindent \textbf{Discrepancy between Simulated and Observed Motions.}
A rigid body's `state' can be succinctly captured by its 6D pose, linear velocity, and angular velocity. Given a current state $s_{t}$ and the specification of the applied forces, one can compute using a physics simulator $\mathcal{P}$, the resulting state $s_{t+1} \equiv \mathcal{P}(s_t, f_t, C_t)$. Therefore, given the initial state, and predicted forces and contact points, we can simulate the entire trajectory under these forces and obtain a resulting state at each timestep.

One possible way to measure the discrepancy between this resulting simulated motion and the observed one is to penalize the difference between the corresponding 6D poses. However, our annotated `ground-truth' 6D poses are often not accurate (due to partial visibility \etc.), and this measure of error is not robust. Instead, we note that we can directly use the annotated 2D keypoint locations to measure the error, by penalizing the re-projection error between the projected keypoints under the simulated pose and the annotated locations of the observed ones. We define a loss function that penalizes this error:
\begin{equation}
\label{eq:keypoint}
\mathcal{L}_{keypoint}(l^{kp}_t, s_t) := \lVert l^{kp}_t - \pi(R_t, T_t)  \rVert^2
\end{equation}
Here, $l^{kp}_t$ is the annotated 2D location for keypoints, $\pi$ is the projection operator which transforms 3D keypoints on the model to the camera frame under the (simulated) rotation and the translation at $s_t = (R_t, T_t)$.

\vspace{1mm}
\noindent \textbf{Differentiable Physics Simulation.} 
To allow learning using the objective above, we require a differentiable physics simulator $\mathcal{P}$. While typical general-purpose simulators are unfortunately not differentiable, we note that the number of input variables (state $s_t$, forces $f_t$, and contact points $C_t$) in our scenario is low-dimensional. We can therefore use finite difference method to calculate the gradients of the outputs of the simulation with respect to its inputs.

In order to calculate the derivative of output with respect to input, we need to calculate the partial derivatives $\frac{\partial S_{t+1}}{S_t}$, $\frac{\partial S_{t+1}}{f_t}$ and $\frac{\partial S_{t+1}}{C}$. We use the approximation
\begin{equation}
\label{eq:finite_diff}
\frac{df}{dx}\cong\frac{f(x+h) - f(x-h)}{2h} 
\end{equation}
where $h$ is a small constant. As $s_t \in \mathbb{R}^{13}$, $f_t \in \mathbb{R}^{k\times3}$, and $C_t \in  \mathbb{R}^{k\times 3}$ ($k=5$ is the number of contact points), we can compute the gradients w.r.t the input using  $(13 + 3k + 3k) + 1$ calls to the simulator $\mathcal{P}$ (the last call is for calculating $f(x)$). We use the PyBullet simulator~\cite{pybullet} for our work, and find that each (differentiable) call takes only 0.12 seconds.

\subsection{Putting it together: Joint Learning of Forces and Contact Points}
\seclabel{learning}

\begin{figure}[tp]
    \centering
    \includegraphics[width=20pc]{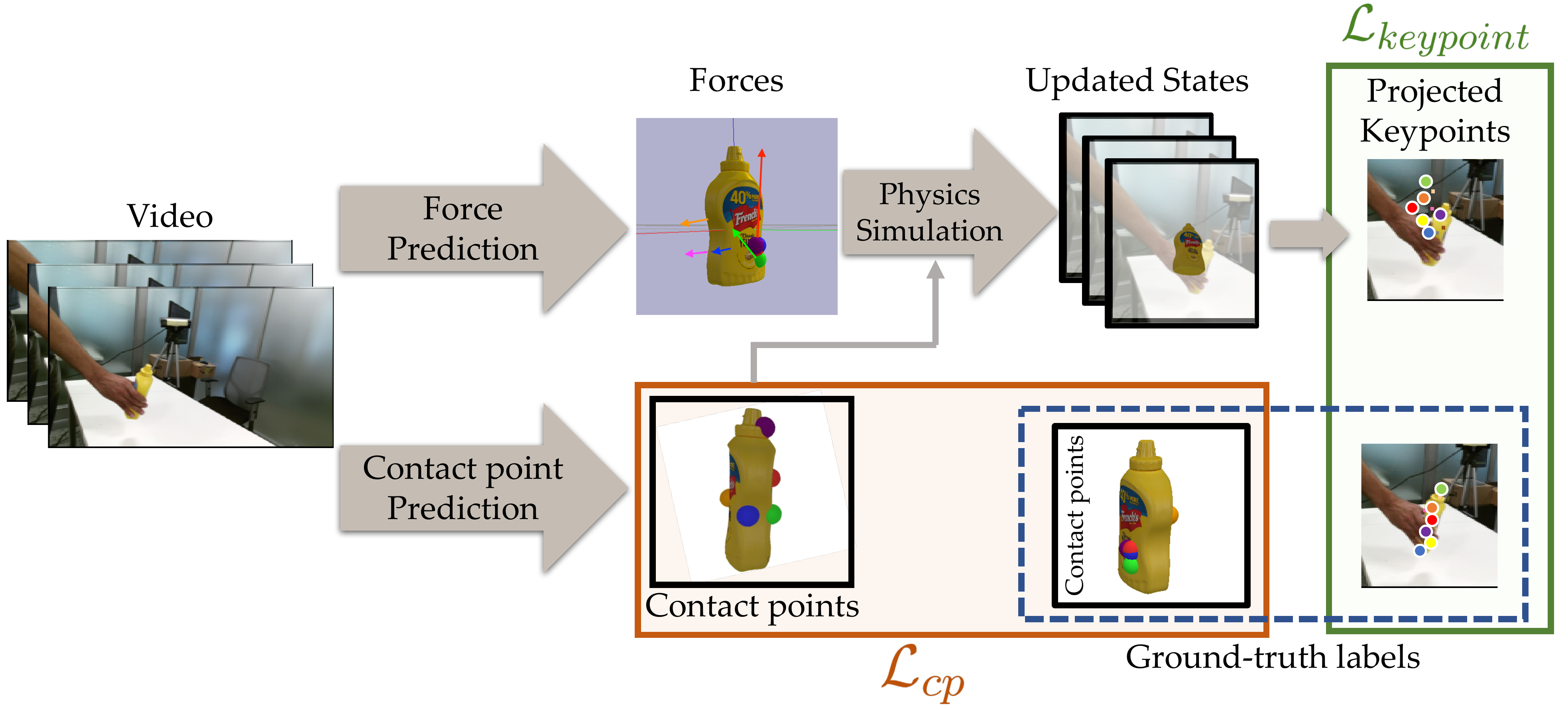}
    \caption{\textbf{Training schema.} Given a video as input, the model predicts the forces and their corresponding contact points. We then apply these forces on the object mesh in physics simulation and jointly optimize for keypoint projection loss $\mathcal{L}_{keypoint}$ and contact point prediction loss $\mathcal{L}_{cp}$ with the aim to imitate the motion observed in the video.}
    \label{fig:objectives}
    \vspace{-1em}
\end{figure}

\begin{figure*}[tp]
    \centering
    \includegraphics[width=40pc]{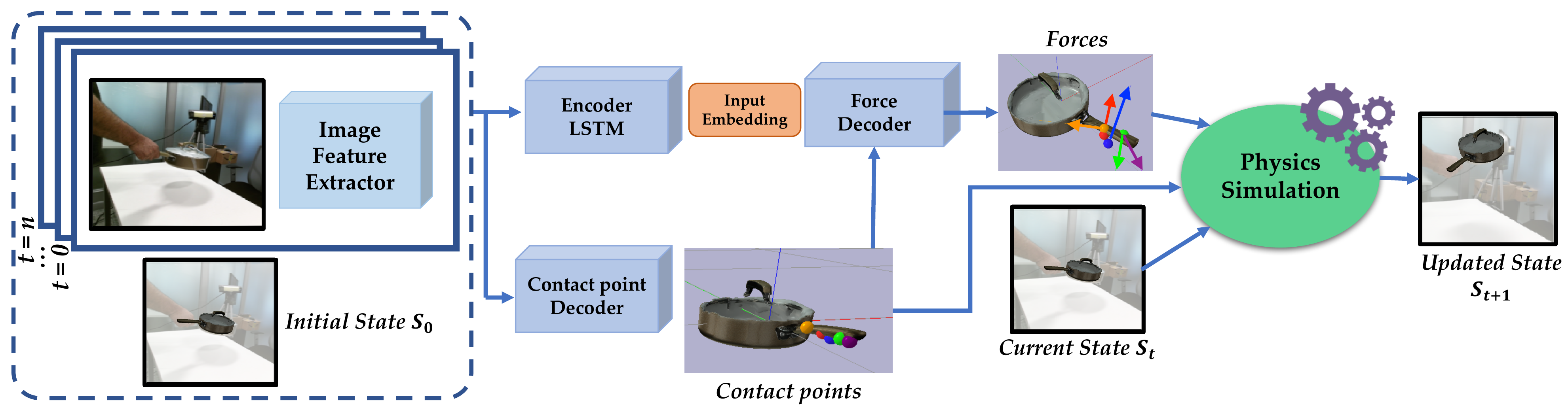}
    \caption{\textbf{Model overview.} Given a video of a person moving an object, along with the initial pose of the object, our network predicts the human contact points and the forces applied at those for each time step. The implied effects of these forces can then be recovered by applying them in a physics simulation. Using the gradients through this simulated interaction, our model learns how to optimize for its two objectives: minimizing the error in projection of the object to camera frame, and predicting the accurate contact points.}
    \label{fig:model}
    \vspace{-1.5em}
\end{figure*}

Given a video and the initial state of the object, we encode the sequence into a visual embedding, and use this embedding to predict the contact points and their corresponding forces. We then apply these forces in the physics simulation to infer the updated state of the object, and use it in addition to the sequence embedding to iteratively predict the subsequent forces to be applied. This can help the network to adapt to the possible mistakes it might have made, and change the forces in the next steps accordingly (Figure ~\ref{fig:model}).

To train our model we have two objectives: (1) to minimize the keypoint re-projection error, that help with reducing the discrepancy between the object trajectory in simulation and the one seen in the video (Equation~\ref{eq:keypoint}), and (2) to minimize the error in contact point prediction in compare to the ground-truth (Figure~\ref{fig:objectives}). The objective we use for optimizing the contact point estimation is defined as,
\begin{equation}
\label{eq:cp}
   \mathcal{L}_{cp}(C_t, \hat{C_t}) := \sum_{i=0,\dots,k}{\lVert c_i - \hat{c}_i \rVert^2},
\end{equation}
where $k$ is the number of contact points, and $C_t$ and $\hat{C}_t$ are the ground truth and predicted contact points at time $t$.

We note that the contact point decoder gets supervisory signals both from contact point loss and the keypoint loss. We believe this constrains contact point prediction to generate physically plausible motions as seen in videos. In experiments, we show this joint loss leads to improvement even in contact point prediction.

\vspace{1mm}
\noindent \textbf{Training details.} 
The backbone for obtaining primary image features is ResNet18~\cite{resnet} pre-trained on ImageNet~\cite{imagenet}. We take the features (which are of size 512x7x7) before the average pooling. For all the experiments we use batch size 64, Adam optimizer with learning rate of 0.001, videos of length 10 frames, with frequency of 30 fps, $h_f, h_s=0.01$ and $h_c=0.05$ for approximating gradients (refer to Equation~\ref{eq:finite_diff}) for force, state and contact point respectively.

We calculate the direction of the gravity in world coordinates and use it in physics simulation to ensure realistic behavior.
To train our model we first train each branch in isolation using $\mathcal{L}_{cp}$ for contact point prediction and $\mathcal{L}_{keypoint}$ for force prediction modules, then we jointly optimize for both objectives and train end to end.

\section{Experiments}
The area of physical understanding of human-object interaction is largely unexplored, and there are no established benchmark datasets or evaluation metrics. In this work, we use our novel dataset to provide empirical evaluations, and in particular to show that: a) our results are qualitatively meaningful; and (b) individual components and loss terms are quantitatively meaningful (ablations). We will also demonstrate that a physical understanding of human-object interactions leads to improvement even in individual components such as contact point estimation and estimating object poses; and (c) finally we will demonstrate the generalization power by showing that our network learns a rich representation that can use few examples to generalize to manipulating novel objects.

\vspace{1mm}
\noindent {\bf Evaluation Metric:} The goal of our work is to obtain physical understanding of human-object interactions. But getting ground truth forces is hard, which makes it impossible to quantitatively measure the performance by just force values. So instead of measuring forces, we evaluate if our predicted forces lead to similar motions as depicted in the videos. Therefore, for evaluating our performance, we use the CP and KP error (Equations~\ref{eq:keypoint} and~\ref{eq:cp}) as evaluation metrics. The former measures the $L1$ distance between the predictions and ground truth for contact point (in object coordinates), and the latter measures error in keypoint projection (in image frame). Original image size is $1920\times 1080$ and the keypoint projection error is reported in pixels in the original image dimension. Rotation and translation error are the angular difference (quaternion distance) in rotation and $L2$ distance (meters) in translation respectively.

\subsection{Qualitative Evaluation}
\begin{figure*}[tp]
    \centering
    \includegraphics[width=41pc]{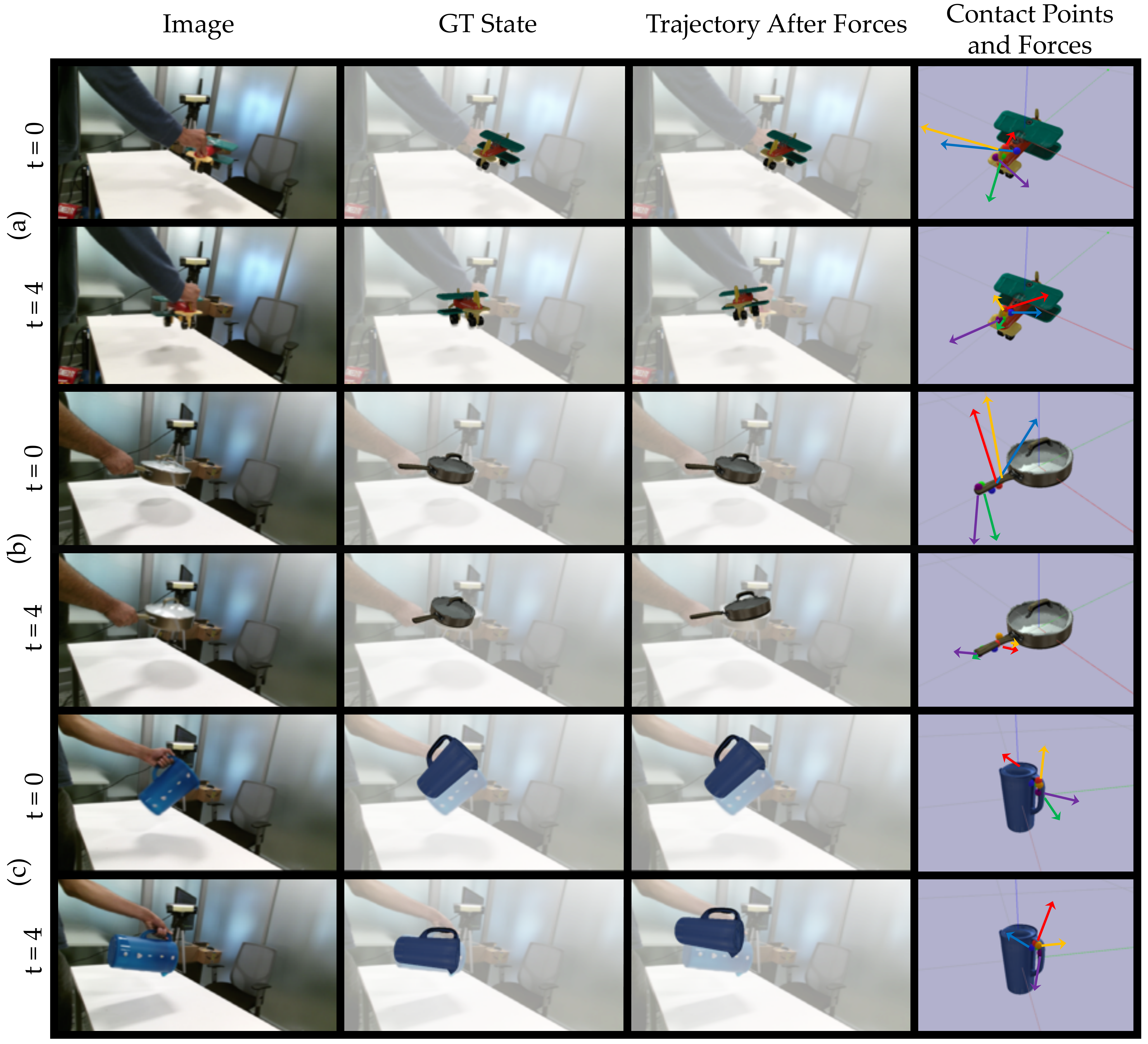}
    \caption{\textbf{Qualitative Results.} We show the results for the model which optimizes for both $\mathcal{L}_{keypoint}$ and $\mathcal{L}_{cp}$. Due to space limitations only Frames for $t=0,4$ are shown. For more videos and contact point visualizations refer to supplementary material.
    }
    \label{fig:qualitative}
    \vspace{-1em}
\end{figure*}

\begin{figure}[tp]
    \centering
    \includegraphics[width=20pc]{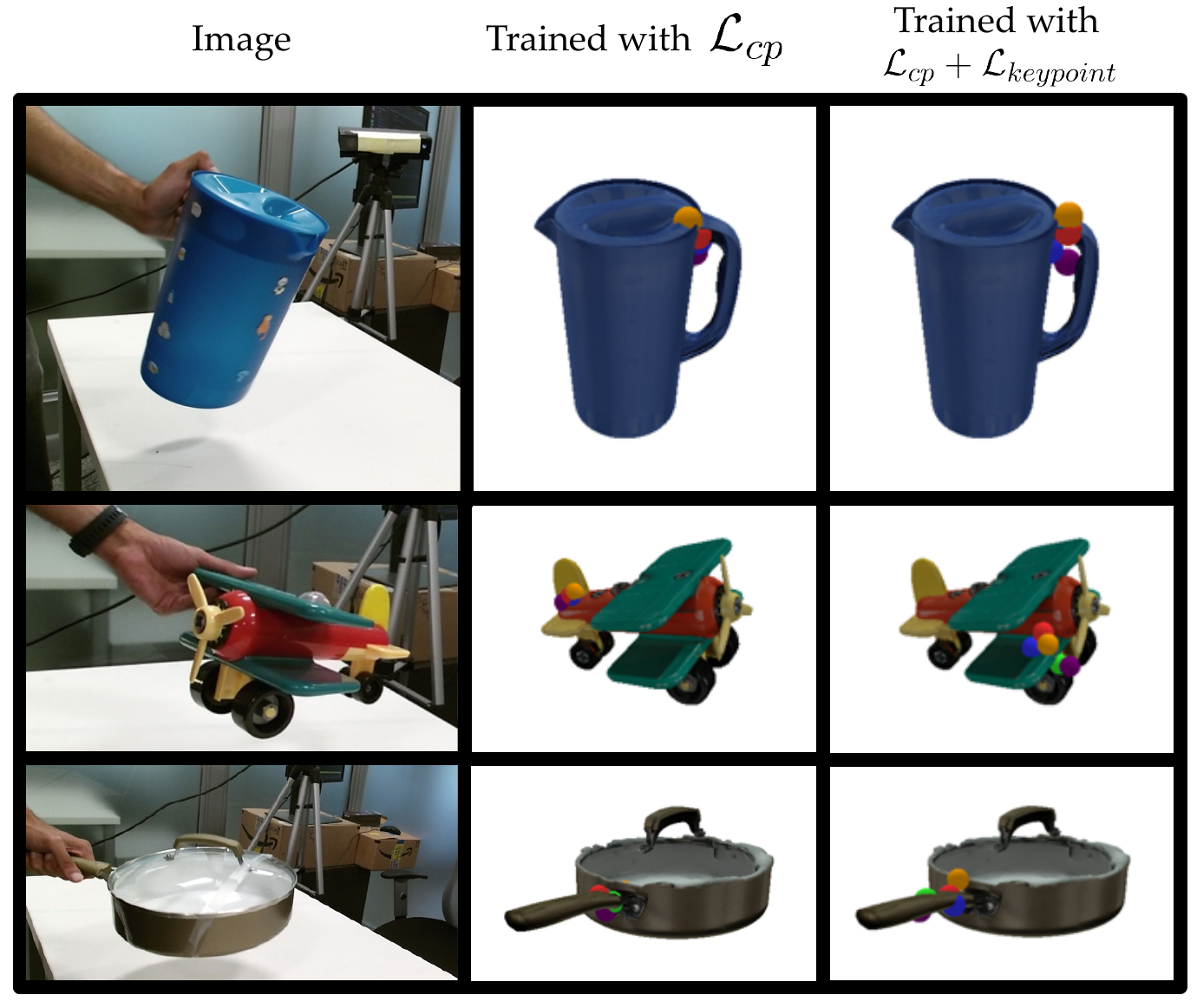}
    \vspace{-5mm}
    \caption{\textbf{Improvements in contact point prediction after joint optimization.} We qualitatively show some examples for which the model makes better contact point predictions when it is trained using both $\mathcal{L}_{keypoint}$ and $\mathcal{L}_{cp}$. Fingers are color-coded (Thumb: orange, Index: Red, Middle: Blue, Ring: Green, Pinky: Purple).}
    \label{fig:cp_improve}
\end{figure}

We first show qualitative results of our full model: joint contact point and force prediction. The qualitative results for force prediction and contact point predictions are shown in Figure~\ref{fig:qualitative} and Figure~\ref{fig:cp_improve} respectively. As the figure shows, our forces are quite meaningful. For example, in case of plane at $t=0$ ( See Figure~\ref{fig:qualitative}(a)), initially the yellow and blue arrow is pushing to left and purple arrow is on the other side is pushing to right. This creates a twist motion and as seen at $t=4$ rotates the plane.  Also, in case of skillet (Figure~\ref{fig:qualitative}(b)), there is a big change in the orientation of the object from $t=0$ to $t=4$, therefore a bigger magnitude of force is required. However, after getting the initial momentum the forces decrease to the minimum needed for maintaining the current state. Refer to project page$^*$ \footnote{$^*$ \url{https://ehsanik.github.io/forcecvpr2020}} for more visualizations.

Next we show few examples of contact point prediction. If contact points are predicted in isolation, small differences can lead to fundamentally different grasps. By enforcing physical meaning to these grasps (via forces), our approach ensures more meaningful predictions. An example is the top of Figure~\ref{fig:qualitative}, where isolated prediction leads to grasp on the rim of the pitcher; but joint prediction leads to prediction of grasp on the handle.

\subsection{Quantitative Evaluation}
To quantitatively evaluate our approach, we measure  performance by evaluating if application of forces as predicted lead to effects as depicted in the videos. Table~\ref{tab:joint_cp_kp} shows the performance in terms of CP metric and KP error. We report both the joint optimization (depicted as $\mathcal{L}_{keypoint} + \mathcal{L}_{cp}$) and the case where contact point prediction and force prediction modules are learned in an isolated manner ($\mathcal{L}_{keypoint} / \mathcal{L}_{cp}$). We observe that the joint model is significantly better (low error), both in contact point prediction and keypoint prediction.

\begin{table*}[t]
\centering
\hfill %
    \begin{tabular}{l c c c c c c}
    \toprule
    Object & Input  & Objective & CP Error &  Keypoint Error (in px) & Rotation Error & Translation Error
    \\
    \midrule 
    Plane & Image & $\mathcal{L}_{keypoint} / \mathcal{L}_{cp}$ &8.79e-2 &109.73&	0.220&	\textbf{0.152}\tabularnewline
	     & Image  & $\mathcal{L}_{keypoint} + \mathcal{L}_{cp}$ & \textbf{7.97e-2} & \textbf{99.89}&\textbf{	0.152}&	0.182	\tabularnewline
	    \midrule 
	    Skillet & Image& $\mathcal{L}_{keypoint} / \mathcal{L}_{cp}$ & 2.37e-2 &70.02&	0.085&	0.094\tabularnewline
	     & Image & $\mathcal{L}_{keypoint} + \mathcal{L}_{cp}$ & \textbf{2.18e-2} & \textbf{65.25}&	\textbf{0.076}&	\textbf{0.063}\tabularnewline
	    \midrule 
	    Pitcher & Image  & $\mathcal{L}_{keypoint} / \mathcal{L}_{cp}$ & 7.3e-2 &131.62&	0.126&	0.212\tabularnewline
	     & Image & $\mathcal{L}_{keypoint} + \mathcal{L}_{cp}$ & \textbf{5.61e-2 }& \textbf{113.40}&	\textbf{0.131}&	\textbf{0.129}\tabularnewline
	    \midrule 
	    Drill & Image & $\mathcal{L}_{keypoint} / \mathcal{L}_{cp}$ & 7.08e-2 &85.86&	0.192&	0.312\tabularnewline
	     & Image & $\mathcal{L}_{keypoint} + \mathcal{L}_{cp}$ & \textbf{6.47e-2} & \textbf{71.00}&\textbf{	0.163}&	\textbf{0.250}\tabularnewline
	    \midrule 
	    \midrule 
	    All objects & Image & $\mathcal{L}_{keypoint} / \mathcal{L}_{cp}$ & 6.83e-2 & 104.57&	0.218&	0.250\tabularnewline
	     & Image & $\mathcal{L}_{keypoint} + \mathcal{L}_{cp}$ & \textbf{6.71e-2} &\textbf{99.38}&	\textbf{0.183}&	\textbf{0.212}\tabularnewline
    \bottomrule
    \end{tabular}
\hfill 
\vspace{2mm}
\vspace{-1em}
\caption{\textbf{Contact point prediction and key point projection error on test set, independent vs. joint optimization} In each set, the first row shows the results of the model optimizing for contact point and keypoint projection error separately, and the second row represents the joint optimization results. End to end training improves the results for both contact point prediction and keypoint projection.}
\label{tab:joint_cp_kp}

\end{table*}

Next, we measure what happens if the contact point prediction was perfectly matched with the annotated ground truth. So, instead of predicting them, we use ground truth contact points.  Given a video, initial pose, and human contact points, we predict the forces applied to each point of contact to replicate the motion. Table~\ref{tab:gt_cp} shows the results for unseen videos. Note that  the training and test sets of contact points are disjoint, so model needs to be able to generalize to applying forces to the object under novel contact point configurations. 

Comparing results in Table~\ref{tab:joint_cp_kp} and Table~\ref{tab:gt_cp} shows that the jointly optimized model ($\mathcal{L}_{keypoint} + \mathcal{L}_{cp}$), (even though it is using the predicted contact points which may have errors), predicts forces that result in better KP metric in compare to the model that uses ground-truth contact points (and is only trained on $\mathcal{L}_{keypoint} $). This shows that jointly reasoning about where to apply the force and what force to apply can help with a better physical understanding.

\begin{table*}[tp]
	\centering
\hfill
	\begin{tabular}{l c c c c c}
	    \toprule
	    Object & Input  & Objective & Keypoint Error (in px) & Rotation Error & Translation Error\tabularnewline
	    \midrule
	    Plane & Image + CP$_{gt}$  & $\mathcal{L}_{keypoint}$ & 104.33&	0.134	&0.199\tabularnewline
	    Skillet & Image + CP$_{gt}$  & $\mathcal{L}_{keypoint}$ &  67.47&	0.075&	0.089	\tabularnewline
	    Pitcher & Image + CP$_{gt}$  & $\mathcal{L}_{keypoint}$ &124.70&	0.130	&0.176	\tabularnewline
	    Drill & Image + CP$_{gt}$  & $\mathcal{L}_{keypoint}$ & 91.25&	0.236	& 0.352	\tabularnewline
	    \midrule 
        \midrule
	    All objects & Image + CP$_{gt}$  & $\mathcal{L}_{keypoint}$ & 102.40&	0.237&	0.273\tabularnewline
	    \bottomrule
	\end{tabular}
\hfill 
\vspace{2mm}
\vspace{-1em}
	\caption{\textbf{Predicting the forces using ground truth contact points} We first train a model to predict the forces on the ground-truth contact points. The first four rows show the quantitative results for training separate model per object using the $\mathcal{L}_{keypoint}$ objective and the last row shows the result for training one shared model for all $8$ objects.}
	\label{tab:gt_cp}
\end{table*}

\noindent{\textbf{Training one model for different objects.}}  We try two training setup for each experiment: (1) training a separate model per object, and (2) sharing weights for one model across all objects (referred to as "All objects"). The common trend of performance improvement for both metrics after joint optimization is observed when training one model for all objects as well. 

\subsection{Few-shot Generalization to Novel Objects}
\begin{figure}[tp]
    \centering
    \includegraphics[width=20pc]{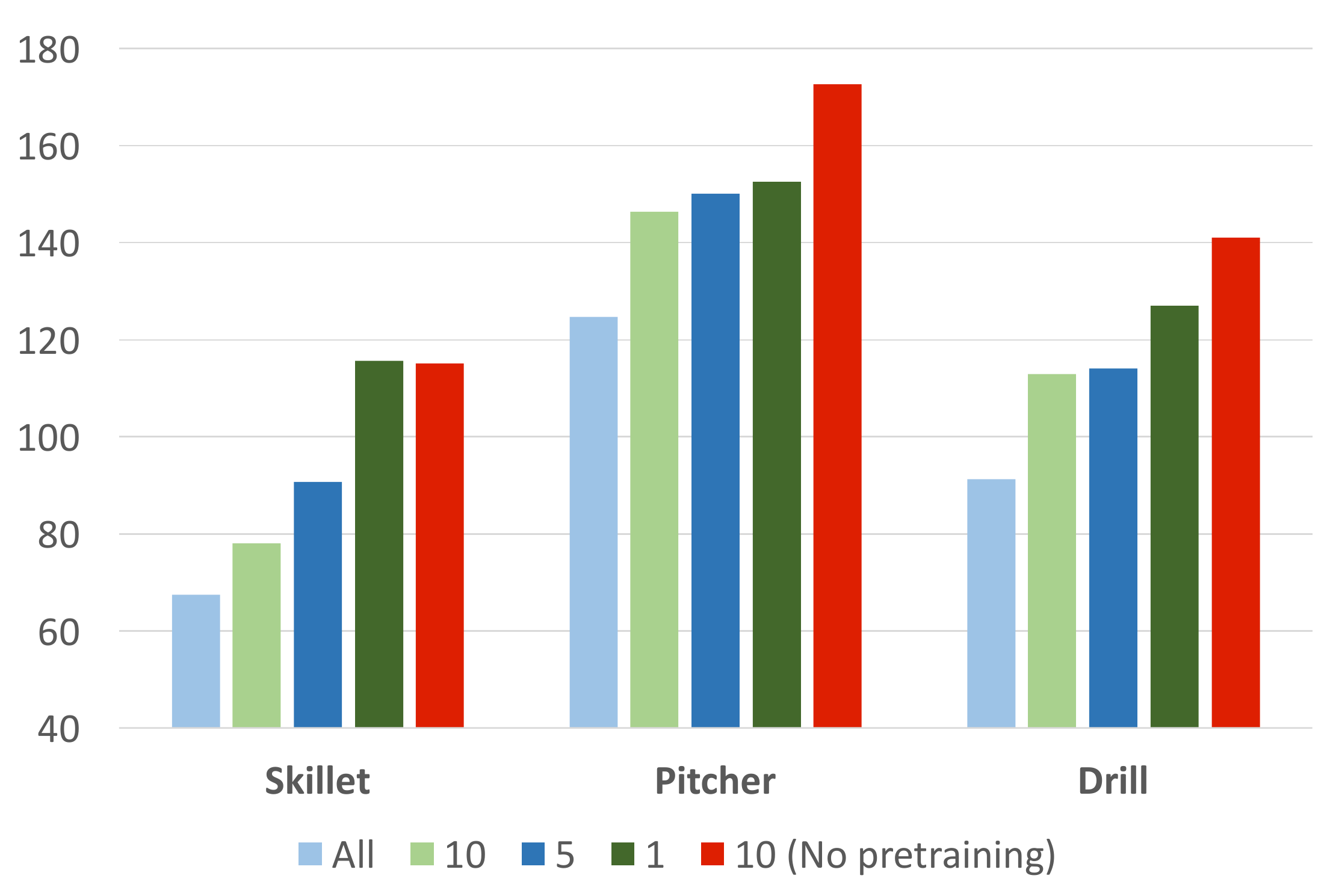}
    \vspace{-1em}
    \caption{\textbf{Few shot experiment.} Showing the keypoint projection error for the model that is trained on planes and fine-tuned on few examples for held out objects (Lower is better). The results are shown for 1, 5, or 10 examples, as well as the model trained on the whole training set. The red bar shows errors for training on 10 examples without plane pre-training.}
    \label{fig:few_shot_plot}
    
\end{figure}

In order to evaluate the representation we learn from this training schema, we train a model for manipulating plane and use that to predict forces on unseen objects using few shot examples. Figure~\ref{fig:few_shot_plot} shows that the representation we learned from predicting forces on one object can generalize to estimating forces for held-out objects using only few examples. Increasing the number of training samples yields better results on the test set. The 10 few shot experiment without pre-training (red) has a significantly lower accuracy than the one with our pre-training (light green) and is comparable with our 1-shot experiment (dark green).

\subsection{Additional Ablation Analysis}

\vspace{1mm}
\noindent \textbf{Regressing the force without simulation gradients.} One alternative approach to solving the force inference problem is to predict the forces without the gradients from simulation. However, this requires having ground-truth labels for forces, which is almost impossible to obtain. So, we try to optimize for a set of pseudo-ground truth forces. 

The goal of this experiment is to investigate how keeping the physics simulation in the training loop can help with understanding physics of the environment and generalizing to unseen trajectories. 

To obtain pseudo-labels,  we optimize a set of valid forces per training example which minimizes the error in keypoint projection. Then we train a model that given the sequence of images and ground truth contact points regresses these forces. The objective is defined as follows:
\[
\mathcal{L}_{force}(f_{gt}, \hat{f}) = \lVert f_{gt} - \hat{f} \rVert^2
\]

Table~\ref{tab:regress_force_ablation} shows that even though the error for training is relatively low, it still fails to generalize to unseen trajectories and gets a high test error. The intuition is that the gradients the network gets from interaction helps with learning a more generalizable representation.

\begin{table}[tp]
	\centering
    \hfill
	\begin{tabular}{l c c c c c}
	    \toprule
	    Set  & Objective & KP (px) & Rotation & Translation\\
	    \midrule
	    Train & $\mathcal{L}_{keypoint}$ & 55.96&	0.129&	0.063\tabularnewline
	    Test  & $\mathcal{L}_{keypoint}$ & 105.60 & 0.138&	0.198\tabularnewline
	    \midrule
	    Train  & $\mathcal{L}_{force}$ &59.86&	0.120&	0.065	\tabularnewline
	    Test & $\mathcal{L}_{force}$ & 185.33 & 0.156 & 0.307 	\tabularnewline
	    \bottomrule
	\end{tabular}
    \hfill

\vspace{2mm}
\vspace{-1em}
	\caption{\textbf{Pseudo-ground truth force regression.} We trained a model to predict the pseudo-ground truth forces for toy airplane from video, initial pose and contact points as input, by directly optimizing for the force prediction ($\mathcal{L}_{force}$). Although the training error is similar to the model trained on keypoint projection loss, it fails to generalize to unseen images and trajectories. }
	\label{tab:regress_force_ablation}
\vspace{-1em}
\end{table}

\vspace{1mm}
\noindent\textbf{Predicting initial state}. We want to evaluate the necessity of giving the initial state as the input to the network. Thus, we try to predict the initial pose instead of using the ground truth. We do so by training a model that given a video as input and contact points in object coordinates (independent of object state), predicts the initial state of the object as well as the forces that are applied on each contact point (Table~\ref{tab:predict_initial}).

\begin{table}[tp]
	\centering
    \hfill
	\begin{tabular}{l c c c}
	    \toprule
	    Input   & KP (px) & Rotation & Translation\\
	    \midrule
	    Image  & 135.80&	0.160&	0.496\tabularnewline
	    Image+$S^{gt}_0$& 128.55&	0.174&	0.329\tabularnewline
	    \bottomrule
	\end{tabular}
    \hfill
\vspace{2mm}
\vspace{-1em}
	\caption{\textbf{Predicting initial state.} The result for the model trained on toy plane object set, predicting the initial pose as well as the forces on ground truth contact points. To see how the error in object initial pose estimation is affecting the performance, we input the ground-truth object pose to the model as input during inference to compare. This model is trained with $\mathcal{L}_{keypoint}$ objective.}
	\label{tab:predict_initial}
	\vspace{-1em}
\end{table}

\begin{figure}[tp]
    \centering
    \includegraphics[width=19pc]{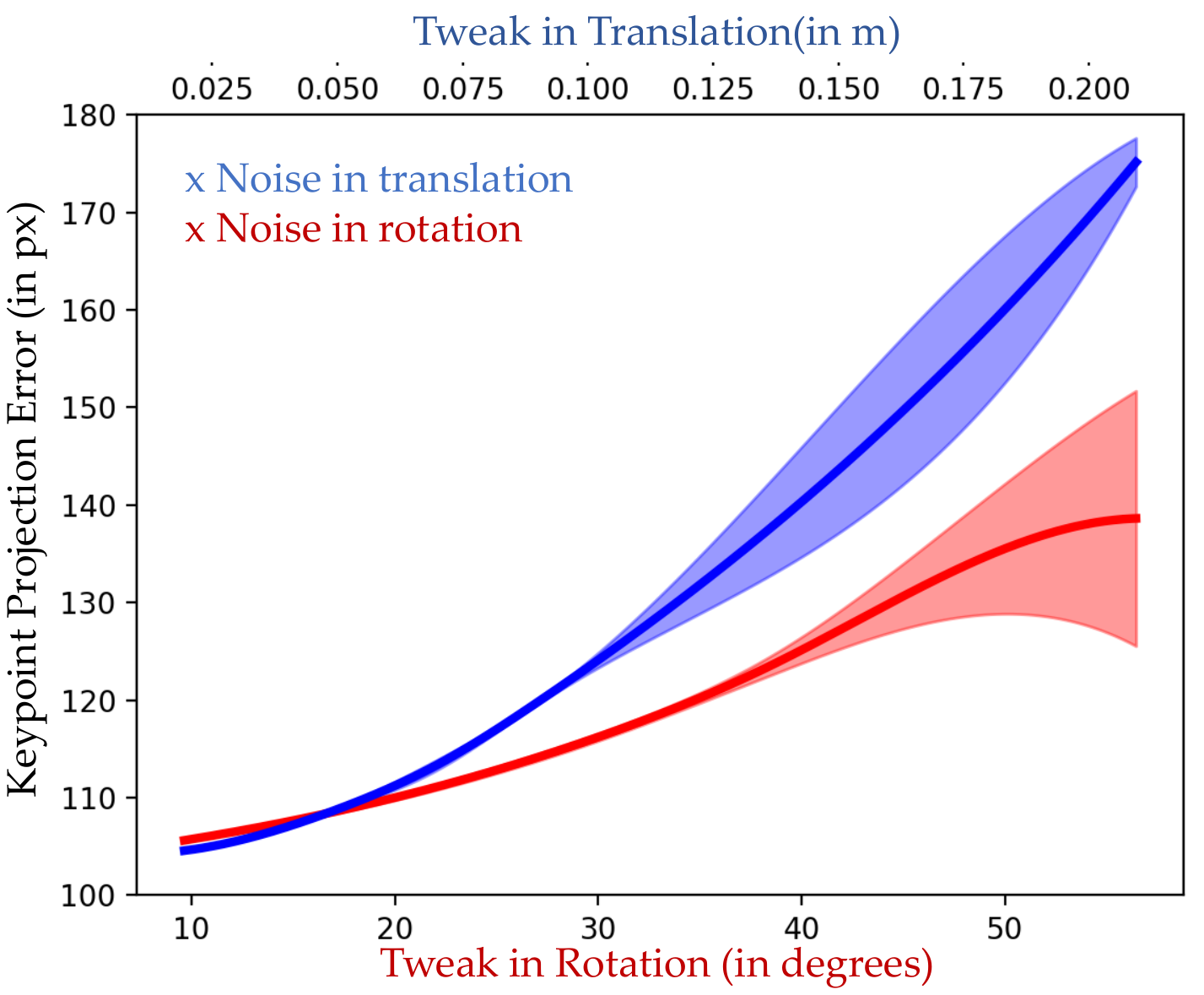}
    \caption{\textbf{Robustness to initial pose.} The changes in the keypoint projection error with respect to the magnitude of the noise added to the initial state. We also calculated the error bars for 5 runs with different random seeds. The labels on the bottom of the chart show the magnitude of the noise in rotation in degrees and the one on the top shows the magnitude of noise in translation in meters.}
    \label{fig:ablation_init}
    \vspace{-1.5em}
\end{figure}

\vspace{1mm}
\noindent \textbf{Adding noise to initial state during inference.} We also want to investigate the effects of adding noise to initial state on performance. This experiment evaluates the robustness of the performance of the model to the initial state estimation. We used the trained model on toy airplane from Table~\ref{tab:gt_cp}, and added noise to the input initial state during inference time. Figure~\ref{fig:ablation_init} shows the changes in KP metric with respect to tweaks in the rotation and translation of the initial state.

\section{Conclusion}
\vspace{-2mm}
We have proposed a model that given a video depicting human-object interaction, predicts the contact points and the corresponding forces such that it replicates the observed motion. We demonstrate that jointly optimizing for both contact point prediction and keypoint projection error improves the results on both tasks in comparison to training models in isolation. We also show that our model learns a meaningful physical representation that can generalize to novel objects using few examples. Since our approach needs textured non-symmetric objects we were able to show these results on 8 objects from the YCB set~\cite{ycb}, but we conjecture that if the keypoint labels in camera frame can be estimated, this method can generalize further than this object set. We believe our work takes step towards integrating action and perception in one common framework -- bringing it closer to real-world robotics. A feasible future work would be to investigate how our model's prediction can speed up the robotic imitation learning procedure.

\small\noindent \textbf{Acknowledgments.} We would like to thank Dhiraj Gandhi for his help with the hardware setup.

{\small
\bibliographystyle{ieee_fullname}
\bibliography{egbib}

\begin{thebibliography}{10}\itemsep=-1pt

\bibitem{akizuki2018tactile}
Shuichi Akizuki and Yoshimitsu Aoki.
\newblock Tactile logging for understanding plausible tool use based on human
  demonstration.
\newblock In {\em British Machine Vision Conference (BMVC)}, 2018.

\bibitem{brahmbhatt2019contactdb}
Samarth Brahmbhatt, Cusuh Ham, Charles~C Kemp, and James Hays.
\newblock Contactdb: Analyzing and predicting grasp contact via thermal
  imaging.
\newblock In {\em Conference on Computer Vision and Pattern Recognition}, pages
  8709--8719, 2019.

\bibitem{ycb}
Berk Calli, Arjun Singh, Aaron Walsman, Siddhartha Srinivasa, Pieter Abbeel,
  and Aaron~M Dollar.
\newblock The ycb object and model set: Towards common benchmarks for
  manipulation research.
\newblock In {\em International Conference on Advanced Robotics (ICAR)}, pages
  510--517. IEEE, 2015.

\bibitem{chebotar2019closing}
Yevgen Chebotar, Ankur Handa, Viktor Makoviychuk, Miles Macklin, Jan Issac,
  Nathan Ratliff, and Dieter Fox.
\newblock Closing the sim-to-real loop: Adapting simulation randomization with
  real world experience.
\newblock In {\em International Conference on Robotics and Automation (ICRA)},
  pages 8973--8979. IEEE, 2019.

\bibitem{pybullet}
Erwin Coumans and Yunfei Bai.
\newblock Pybullet, a python module for physics simulation for games, robotics
  and machine learning.
\newblock \url{http://pybullet.org}, 2016--2019.

\bibitem{de2018end}
Filipe de Avila Belbute-Peres, Kevin Smith, Kelsey Allen, Josh Tenenbaum, and
  J~Zico Kolter.
\newblock End-to-end differentiable physics for learning and control.
\newblock In {\em Advances in Neural Information Processing Systems}, pages
  7178--7189, 2018.

\bibitem{imagenet}
Jia Deng, Wei Dong, Richard Socher, Li-Jia Li, Kai Li, and Li Fei-Fei.
\newblock Imagenet: A large-scale hierarchical image database.
\newblock In {\em Conference on Computer Vision and Pattern Recognition}, pages
  248--255. IEEE, 2009.

\bibitem{doumanoglou2016recovering}
Andreas Doumanoglou, Rigas Kouskouridas, Sotiris Malassiotis, and Tae-Kyun Kim.
\newblock Recovering 6d object pose and predicting next-best-view in the crowd.
\newblock In {\em Conference on Computer Vision and Pattern Recognition}, pages
  3583--3592, 2016.

\bibitem{fragkiadakipredictive}
Katerina Fragkiadaki, Pulkit Agrawal, Sergey Levine, and Jitendra Malik.
\newblock Predictive visual models of physics for playing billiards.
\newblock In {\em International Conference on Learning Representations}, 2016.

\bibitem{garcia2018first}
Guillermo Garcia-Hernando, Shanxin Yuan, Seungryul Baek, and Tae-Kyun Kim.
\newblock First-person hand action benchmark with rgb-d videos and 3d hand pose
  annotations.
\newblock In {\em Conference on Computer Vision and Pattern Recognition}, pages
  409--419, 2018.

\bibitem{gkioxari2018detecting}
Georgia Gkioxari, Ross Girshick, Piotr Doll{\'a}r, and Kaiming He.
\newblock Detecting and recognizing human-object interactions.
\newblock In {\em Conference on Computer Vision and Pattern Recognition}, pages
  8359--8367, 2018.

\bibitem{gupta2015aligning}
Saurabh Gupta, Pablo Arbel{\'a}ez, Ross Girshick, and Jitendra Malik.
\newblock Aligning 3d models to rgb-d images of cluttered scenes.
\newblock In {\em Conference on Computer Vision and Pattern Recognition}, pages
  4731--4740, 2015.

\bibitem{gupta2015visual}
Saurabh Gupta and Jitendra Malik.
\newblock Visual semantic role labeling.
\newblock {\em arXiv preprint arXiv:1505.04474}, 2015.

\bibitem{hamer2010object}
Henning Hamer, Juergen Gall, Thibaut Weise, and Luc Van~Gool.
\newblock An object-dependent hand pose prior from sparse training data.
\newblock In {\em Computer Society Conference on Computer Vision and Pattern
  Recognition}, pages 671--678. IEEE, 2010.

\bibitem{resnet}
Kaiming He, Xiangyu Zhang, Shaoqing Ren, and Jian Sun.
\newblock Deep residual learning for image recognition.
\newblock In {\em Conference on Computer Vision and Pattern Recognition}, pages
  770--778. IEEE, 2016.

\bibitem{hu2019chainqueen}
Yuanming Hu, Jiancheng Liu, Andrew Spielberg, Joshua~B Tenenbaum, William~T
  Freeman, Jiajun Wu, Daniela Rus, and Wojciech Matusik.
\newblock Chainqueen: A real-time differentiable physical simulator for soft
  robotics.
\newblock In {\em International Conference on Robotics and Automation (ICRA)},
  pages 6265--6271. IEEE, 2019.

\bibitem{huang2015we}
De-An Huang, Minghuang Ma, Wei-Chiu Ma, and Kris~M Kitani.
\newblock How do we use our hands? discovering a diverse set of common grasps.
\newblock In {\em Conference on Computer Vision and Pattern Recognition}, pages
  666--675, 2015.

\bibitem{hwang2017inferring}
Wonjun Hwang and Soo-Chul Lim.
\newblock Inferring interaction force from visual information without using
  physical force sensors.
\newblock {\em Sensors}, 17(11):2455, 2017.

\bibitem{kehl2016deep}
Wadim Kehl, Fausto Milletari, Federico Tombari, Slobodan Ilic, and Nassir
  Navab.
\newblock Deep learning of local rgb-d patches for 3d object detection and 6d
  pose estimation.
\newblock In {\em European Conference on Computer Vision}, pages 205--220.
  Springer, 2016.

\bibitem{li2018deepim}
Yi Li, Gu Wang, Xiangyang Ji, Yu Xiang, and Dieter Fox.
\newblock Deepim: Deep iterative matching for 6d pose estimation.
\newblock In {\em European Conference on Computer Vision (ECCV)}, pages
  683--698, 2018.

\bibitem{li2019estimating}
Zongmian Li, Jiri Sedlar, Justin Carpentier, Ivan Laptev, Nicolas Mansard, and
  Josef Sivic.
\newblock Estimating 3d motion and forces of person-object interactions from
  monocular video.
\newblock In {\em Conference on Computer Vision and Pattern Recognition}, pages
  8640--8649, 2019.

\bibitem{liu2008design}
Tao Liu, Yoshio Inoue, and Kyoko Shibata.
\newblock Design of low-cost tactile force sensor for 3d force scan.
\newblock In {\em Sensors}, pages 1513--1516. IEEE, 2008.

\bibitem{mottaghi2016newtonian}
Roozbeh Mottaghi, Hessam Bagherinezhad, Mohammad Rastegari, and Ali Farhadi.
\newblock Newtonian scene understanding: Unfolding the dynamics of objects in
  static images.
\newblock In {\em Conference on Computer Vision and Pattern Recognition}, pages
  3521--3529, 2016.

\bibitem{mottaghi2016happens}
Roozbeh Mottaghi, Mohammad Rastegari, Abhinav Gupta, and Ali Farhadi.
\newblock “what happens if...” learning to predict the effect of forces in
  images.
\newblock In {\em European Conference on Computer Vision}, pages 269--285.
  Springer, 2016.

\bibitem{pham2015towards}
Tu-Hoa Pham, Abderrahmane Kheddar, Ammar Qammaz, and Antonis~A Argyros.
\newblock Towards force sensing from vision: Observing hand-object interactions
  to infer manipulation forces.
\newblock In {\em Conference on Computer Vision and Pattern Recognition}, pages
  2810--2819, 2015.

\bibitem{pham2017hand}
Tu-Hoa Pham, Nikolaos Kyriazis, Antonis~A Argyros, and Abderrahmane Kheddar.
\newblock Hand-object contact force estimation from markerless visual tracking.
\newblock {\em Transactions on Pattern Analysis and Machine Intelligence},
  40(12):2883--2896, 2017.

\bibitem{purushwalkam2018bounce}
Senthil Purushwalkam, Abhinav Gupta, Danny Kaufman, and Bryan Russell.
\newblock Bounce and learn: Modeling scene dynamics with real-world bounces.
\newblock In {\em International Conference on Learning Representations}, 2019.

\bibitem{sahin2018category}
Caner Sahin and Tae-Kyun Kim.
\newblock Category-level 6d object pose recovery in depth images.
\newblock In {\em European Conference on Computer Vision (ECCV)}, 2018.

\bibitem{song2014sliding}
Shuran Song and Jianxiong Xiao.
\newblock Sliding shapes for 3d object detection in depth images.
\newblock In {\em European conference on computer vision}, pages 634--651.
  Springer, 2014.

\bibitem{tremblay2018deep}
Jonathan Tremblay, Thang To, Balakumar Sundaralingam, Yu Xiang, Dieter Fox, and
  Stan Birchfield.
\newblock Deep object pose estimation for semantic robotic grasping of
  household objects.
\newblock In {\em Conference on Robot Learning}, pages 306--316, 2018.

\bibitem{wang2019learning}
He Wang, S{\"o}ren Pirk, Ersin Yumer, Vladimir~G Kim, Ozan Sener, Srinath
  Sridhar, and Leonidas~J Guibas.
\newblock Learning a generative model for multi-step human-object interactions
  from videos.
\newblock In {\em Computer Graphics Forum}, volume~38, pages 367--378. Wiley
  Online Library, 2019.

\bibitem{wu2015galileo}
Jiajun Wu, Ilker Yildirim, Joseph~J Lim, Bill Freeman, and Josh Tenenbaum.
\newblock Galileo: Perceiving physical object properties by integrating a
  physics engine with deep learning.
\newblock In {\em Advances in Neural Information Processing Systems}, pages
  127--135, 2015.

\bibitem{xiang2017posecnn}
Yu Xiang, Tanner Schmidt, Venkatraman Narayanan, and Dieter Fox.
\newblock Posecnn: A convolutional neural network for 6d object pose estimation
  in cluttered scenes.
\newblock In {\em Robotics: Science and Systems (RSS)}, 2018.

\bibitem{yatskar2016situation}
Mark Yatskar, Luke Zettlemoyer, and Ali Farhadi.
\newblock Situation recognition: Visual semantic role labeling for image
  understanding.
\newblock In {\em IEEE Conference on Computer Vision and Pattern Recognition},
  pages 5534--5542, 2016.

\end{thebibliography}
}

\twocolumn[{%
\renewcommand\twocolumn[1][]{#1}%
\vspace{-3em}
\maketitle
\vspace{-2em}
\begin{center}
   \centering \includegraphics[width=\textwidth]{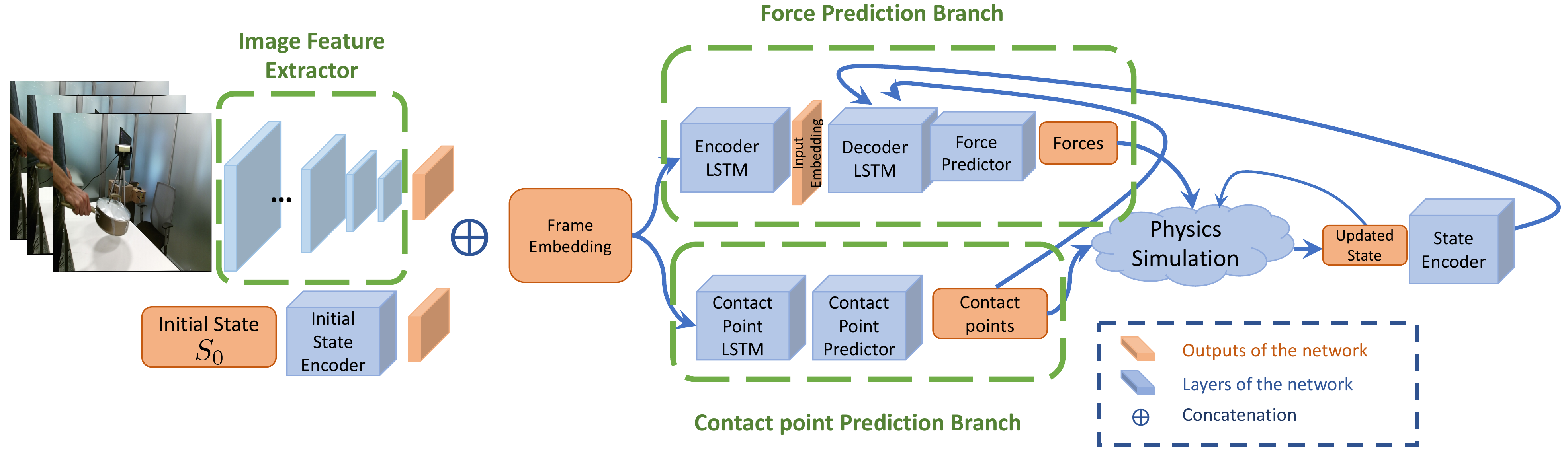} \captionof{figure}{\textbf{Architecture details.} }
   \label{fig:complete_model}
\end{center}%
}]

\section{Supplementary Material}

\begin{figure*}[tp]
    \centering
    \includegraphics[width=30pc]{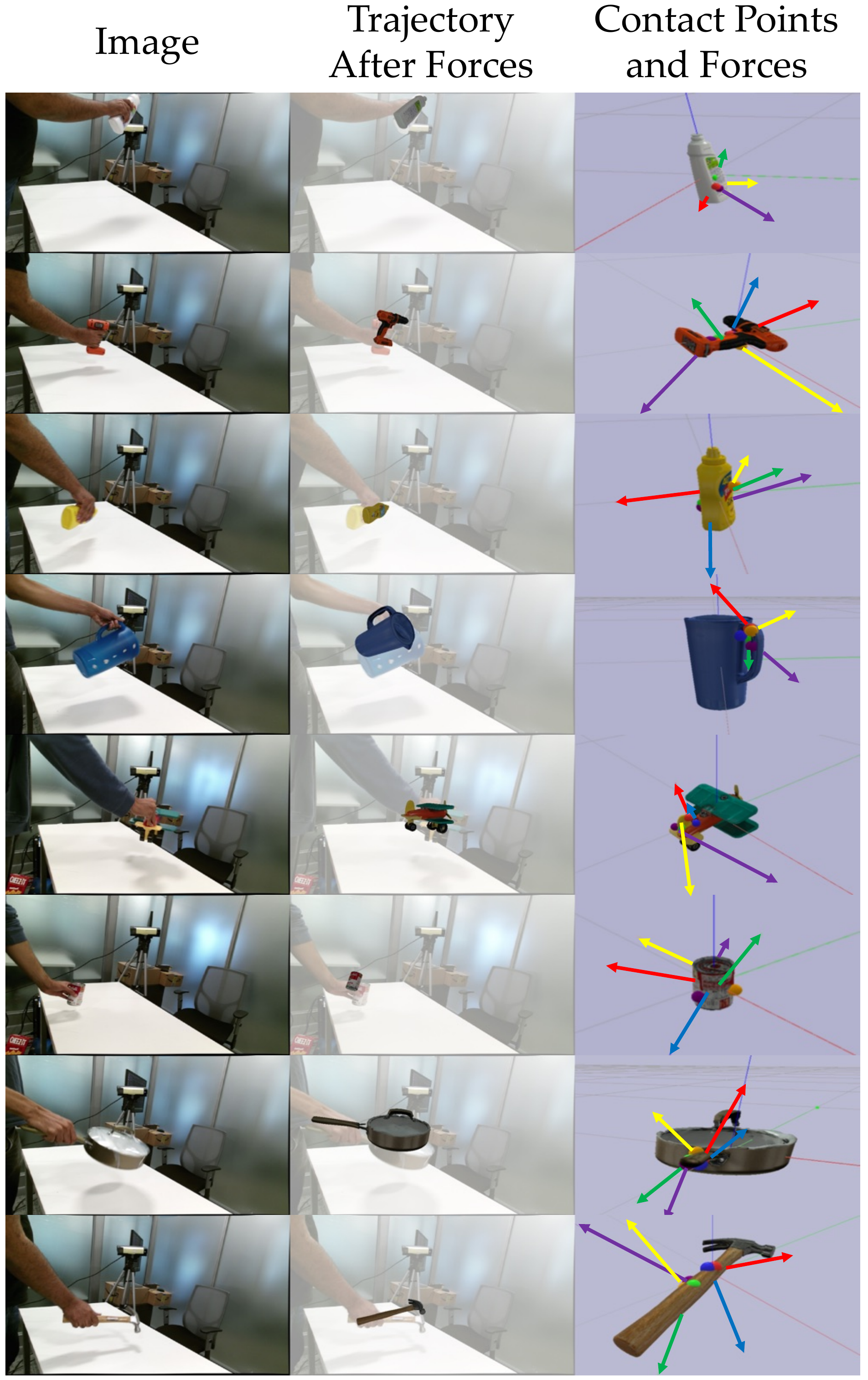}
    \caption{\textbf{More qualitative frames.} We show the estimated contact points and forces on variety of objects. For more videos and contact point visualizations refer to project's webpage (\url{https://ehsanik.github.io/forcecvpr2020}).}
    \label{fig:suppl_qual}
\end{figure*}

We first explain the details of the data collection, and data pre-processing. Then we continue by discussing the architecture and hyper-parameters in detail to navigate the researchers to reproduce the results. Code in PyTorch and the dataset will be released to the public later. Finally, we illustrate one frame per object showing the simulated trajectory, the contact points and the forces applied to them at a specific time step (Figure~\ref{fig:suppl_qual}). Code, data and the video of qualitative results are available in \url{https://ehsanik.github.io/forcecvpr2020}.

\subsection{Dataset}
\label{sec:dataset_suppl}

\vspace{1mm}
\noindent \textbf{Annotations for Motion Estimation.}
We manually defined 10 keypoints per object and asked Turkers to annotate these points on each video frame if the corresponding keypoint was visible. Given the known locations of these points of the object mesh, and their (annotated) projections on the image, we can recover the object's 6D pose w.r.t. the camera using the PnP algorithm. However, we note that due to noisy annotations, partial visibility, or co-planarity of points, the recovered poses are often inaccurate. 

\vspace{1mm}
\noindent \textbf{Annotating Contact Points.}
Instead of directly marking contact points for each interaction on the object mesh, we note that finding their projections on the image \ie pixel locations of fingers is more natural for annotators. Given the annotations of their projections across the interaction frames, we can find the contact points on the mesh surface that project at these locations under the poses $(R_t, T_t)$ inferred in the previous step.
This is achieved by solving an optimization problem of finding $C$ for each interaction sequence such that,
\begin{equation}
   {arg\,min}_C \sum_t \lVert \pi (C, R_t, T_t) - l^c_t \rVert^2,
\end{equation}

where $t=0,\dots, n$ is the time step, $C\in \mathbb{R}^{k\times 3}$ is the set of contact points (k is the number of contact points), $R_t$ and $T_t$ are the rotation and translation of the object in world coordinate, respectively, $l^c_t$ is the pixel annotation for contact points from Mechanical Turk, and $\pi$ is the projection of the 3D points from the world coordinate to the camera frame.  Note that the camera is static and does not move throughout each interaction sequence.

Detailed statistics of the dataset is available in Table~\ref{tab:data_stat}.

\begin{table}[tp]
    \centering
    \begin{tabular}{|c|c| c|c|}
        \hline
        Object & Train  & Test & Validation \\
        \hline
        Pitcher    &843/12    &135/4&    219/3\\
        Bleach    &994/15    &298/5&    217/3\\
        Skillet    &904/12    &336/4&    397/4\\
        Drill    &1202/14    &351/4&    344/5\\
        Hammer    &970/13    &297/4    &302/4\\
        Plane    &1762/19&    444/6&    339/6\\
        Tomato soup    &634/12&    224/3&    352/3\\
        Mustard    &1007/14    &299/5&    384/4\\
        \hline
        Total    &8316/111    &2384/31&    2554/32\\
        \hline
    \end{tabular}
    \caption{\textbf{Dataset Statistics} Distribution of number of frames and number of distinct contact points per split. (\#frames/\# distinct contact points)}
    \label{tab:data_stat}
\end{table}

\subsection{Architecture Details}

The initial state encoder, contact point predictor and force predictor in Figure~\ref{fig:complete_model} are each three fully connected layers. Image feature extractor is a Resnet18 without the last fully connected and average pooling layer followed by a point-wise convolution. Encoder LSTM and contact point LSTM are each a unidirectional Long-Short Term Memory with 3 layers of hidden size 512 and Decoder LSTM is a single LSTM cell with hidden size 512. The inputs to Decoder LSTM at each time step $t$, are the encoding for the current state $S_t$, the predicted contact points $C_t$, the input embedding (which is the embedding of the entire sequence), and the frame embedding for time $t+1$.  State encoder is three fully connected layers that embed current rotation and translation of the object as well as its linear and angular velocity.

\subsection{Training Details}

To avoid over-fitting and enhance generalization, we randomly jitter the hue, saturation, and brightness of the images by $0.05$. We train each one of our models until convergence on the training set, which takes between 30-60 epochs. Training each epoch takes 18-35 minutes on one GPU and 12 core CPU. We use a learning rate of $0.0001$ for few shot experiments and $0.001$ for the rest. We resize the input images to $224\times 224$ before giving them as input to the feature extraction block.

\vspace{3mm}

\textbf{Qualitative results follow on the next page.
}

\end{document}